\documentclass[sigconf,nonacm]{acmart}
\makeatletter
\def\@ACM@checkaffil{
    \if@ACM@instpresent\else
    \ClassWarningNoLine{\@classname}{No institution present for an affiliation}%
    \fi
    \if@ACM@citypresent\else
    \ClassWarningNoLine{\@classname}{No city present for an affiliation}%
    \fi
    \if@ACM@countrypresent\else
        \ClassWarningNoLine{\@classname}{No country present for an affiliation}%
    \fi
}
\makeatother
\usepackage{hyperref}
\usepackage{url}
\usepackage{booktabs}
\usepackage{amsfonts}
\usepackage{nicefrac}
\usepackage{microtype}
\usepackage{xcolor}
\usepackage{graphicx}
\usepackage{colortbl}
\usepackage{graphicx}
\usepackage{amsmath}

\usepackage{amssymb}
\usepackage{booktabs}
\usepackage{epsfig}
\usepackage{multirow}
\usepackage{makecell}
\usepackage{float}
\usepackage{subcaption}
\usepackage{pifont}
\usepackage{enumitem}

\AtBeginDocument{%
  }

\begin{document}

\title{MobileVidFactory: Automatic Diffusion-Based Social Media Video Generation for Mobile Devices from Text}

\author{Junchen Zhu$^1$,\hspace{0.8em} Huan Yang$^2$*,\hspace{0.8em} Wenjing Wang$^2$,\hspace{0.8em} Huiguo He$^2$,\hspace{0.8em} Zixi Tuo$^2$,\hspace{0.8em} Yongsheng Yu$^3$, \hspace{0.8em} Wen-Huang Cheng$^4$,\hspace{0.8em} Lianli Gao$^1$,\hspace{0.8em} Jingkuan Song$^1$*,\hspace{0.8em} Jianlong Fu$^2$,\hspace{0.8em} Jiebo Luo$^3$}
\affiliation{%
 \vspace{0em}\institution{$^1$Center for Future Media, University of Electronic Science and Technology of China \hspace{0.3em} \\$^2$Microsoft Research \hspace{0.3em} $^3$University of Rochester \hspace{0.3em} $^4$National Taiwan University}
}
\affiliation{%
  \institution{junchen.zhu@hotmail.com \hspace{0.1em} \{huayan,v-wenjiwang,v-huiguohe,v-zixituo,jianf\}@microsoft.com \hspace{0.1em}yyu90@ur.rochester.edu}
}
\affiliation{%
  \institution{wenhuang@csie.ntu.edu.tw \hspace{0.1em} lianli.gao@uestc.edu.cn \hspace{0.1em} jingkuan.song@gmail.com \hspace{0.1em} jluo@cs.rochester.edu}
}

\renewcommand{\shortauthors}{Zhu et al.}

\begin{abstract}
Videos for mobile devices become the most popular access to share and acquire information recently. For the convenience of users' creation, in this paper, we present a system, namely \textbf{MobileVidFactory}, to automatically generate vertical mobile videos where users only need to give simple texts mainly. Our system consists of two parts: basic and customized generation. In the basic generation, we take advantage of the pretrained image diffusion model, and adapt it to a high-quality open-domain vertical video generator for mobile devices. As for the audio, by retrieving from our big database, our system matches a suitable background sound for the video. Additionally to produce customized content, our system allows users to add specified screen texts to the video for enriching visual expression, and specify texts for automatic reading with optional voices as they like.
\end{abstract}


\maketitle

\begin{figure}[t]
    \centering
    \includegraphics[width=1.0\linewidth]{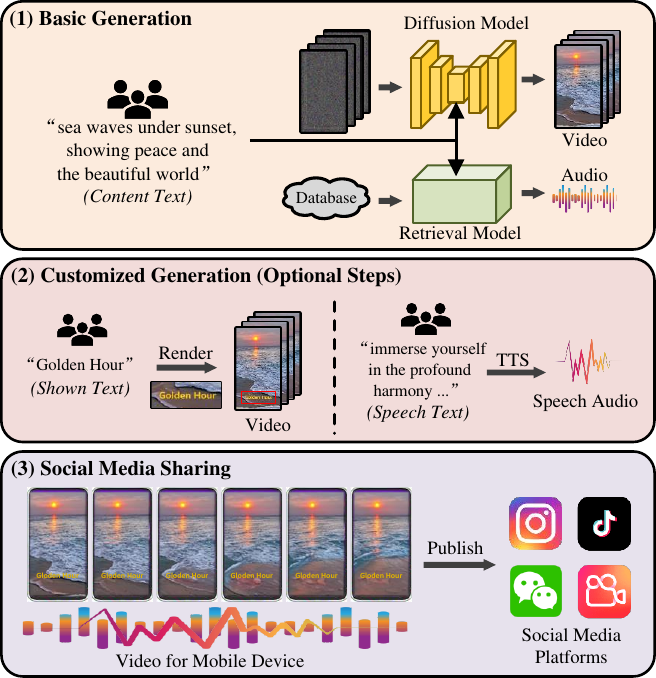}
    \caption{Illustration of our system. Given an input text, our model generates the basic visual content and retrieves the audio part for the video. Then, users can further add visible texts to the decorate visual part or human speech to enrich the expression with texts as well.}
    \label{fig:framework}
\end{figure}

\let\thefootnote\relax\footnote{*Huan Yang and Jingkuan Song are the corresponding authors.}
\section{Introduction}
The rise of mobile devices, e.g., smartphones, and the exponential growth in social media has revolutionized the way we consume video content. A prominent shift in viewing habits has occurred, with vertical videos gaining significant traction among mobile users. This paper investigates the landscape of vertical mobile video generation, focusing on the distinct challenges, opportunities, and implications it presents for content creators and consumers alike. By exploring this evolving form of visual storytelling, we uncover the transformative potential it holds for the future of content consumption. However, creating such videos is not easy for everyone, as professional skills are always required from shooting to editing.

In this paper, we present a novel system \textbf{MobileVidFactory} to elegantly help users create their own videos for mobile devices. 
In summary, our system is leading in the following aspects:
\begin{itemize}
    \item[1)]
    Our system is the first automatic mobile-device video generation framework to the best of our knowledge, which can be used to generate high-quality, sounding, and vertical videos with simple text from users.
    \item[2)]
    Our system contains both deep networks to generate the basic visual and auditory content, and fixed modules to add customized content for users, where all components are controlled by text mainly. This makes our system meet different needs of users as much as possible and operate easily.
    \item[3)] 
    Our system is able to learn to synthesize open-domain high-quality vertical videos after finetuning with a few vertical video data, as we design an efficient training strategy to learn motion information from large-scale horizontal video datasets and migrate it losslessly to vertical video.
\end{itemize}

\section{System}
We aim to automatically generate vertical videos that can be directly used on mobile devices and can be easily controlled by simple texts from users.
As shown in Fig.\ref{fig:framework}, our system contains two generation processes: basic generation, which produces the elemental video and audio, and customized generation, which offers the option to add screen text and dubbing.

\subsection{Basic Generation}
In the basic generation, not like prior works~\cite{MMDiffusion,svgvqgan}, we independently produce the visual and auditory content. The video is synthesized using a diffusion model, and the audio is obtained through a retrieval model.
To generate high-quality vertical videos, we pretrain our model on a large-scale horizontal video dataset~\cite{webvid,VideoFactory}, and adapt it to the vertical generation using a few vertical data. Specifically, we firstly extend a pretrained image diffusion model~\cite{latentdiffusion,stablediffusion} to fit the spatial distribution (e.g., new aspect ratio and picture style) of the video dataset by adding learnable blocks in each level of the U-Net~\cite{unet} network, and only train these blocks while fixing all pretrained layers. 
Then, building upon previous works~\cite{MakeAVideo,VideoLDM}, we add new temporal layers, including temporal 1D residual blocks and transformer blocks~\cite{transformer}, and utilize the video dataset to train these additional layers~\cite{moviefactory}. 
After pretraining, we fix all layers except for the ones added to adjust the spatial distribution. We then finetune these spatial blocks to adapt to the vertical screens of mobile devices using some collected data. 
Additionally, we provide a frame interpolation module~\cite{AMT,TTVFI} to achieve smoother motion in certain cases.
To match vivid audio to the visual content, we adopt the previous work~\cite{OncescuKHAA21,Koepke21} as our retrieval model.
The network employs a text encoder and an audio encoder to encode text and audio data, respectively. It then evaluates the distance between the encoded features of multiple pairs to search for the most relevant sample. This retrieval model is trained using a contrastive ranking loss.
In our scenario, we directly use the textual content provided by users to retrieve the top 3 matching audios from the database. Users can then choose their favorite audio and select the most appropriate section to add to the video.

\subsection{Customized Generation}
In our system, we design two optional customized functions for users.
First, we support adding customized screen text to the video, as the addition of text to short mobile videos can provide context, clarify concepts, and enhance viewer engagement.
Text overlays also promote accessibility for individuals with hearing impairments and cater to diverse audiences. 
With the provided text to be shown, the specified font size and font color to use, and the designated position to place, our system directly overlays the text on top of the video for rendering.
Second, to add a personal touch, enhance storytelling, provide explanations, and foster a sense of authenticity, our system allows users to add dubbing to their videos. 
Using sentences from users, we utilize bark~\cite{bark} to transform the text into dubbing, known as text-to-speech (TTS). 
In this process, users can choose the voice they prefer, and various languages, such as English and Chinese, are well supported.

\begin{figure}[t]
    \centering
    \includegraphics[width=\linewidth]{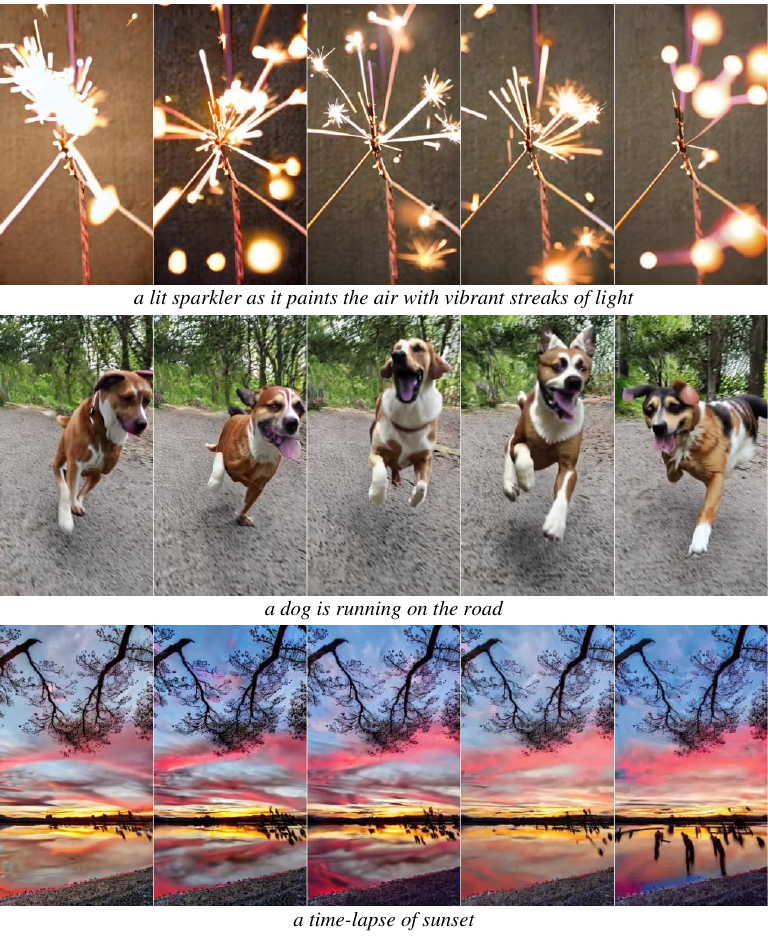}
    \caption{Generated video samples of our system. More samples are included in our video.}
    \label{fig:result}
\end{figure}
\subsection{Demonstration}
We demonstrate some generated videos in Fig.\ref{fig:result}. Focusing on the vertical videos for mobile devices, our system can synthesize frames with abundant details and create compositions that highlight the subject. Additionally, smooth motion can be captured to describe vivid scenes. More samples are included in our video.

\section{Conclusion}
In this paper, we present MobileVidFactory, a system that automatically generates vertical videos for mobile devices using text inputs. 
The system consists of a visual generator that creates high-quality videos by leveraging a pretrained image diffusion model and a finetuning process. 
Users can enrich visual expression by adding specified texts. The audio generator matches suitable background sounds from a database and provides optional text-to-speech narration.

\bibliographystyle{ACM-Reference-Format}
\bibliography{main}


\begin{thebibliography}{16}


\ifx \showCODEN    \undefined \def \showCODEN     #1{\unskip}     \fi
\ifx \showDOI      \undefined \def \showDOI       #1{#1}\fi
\ifx \showISBNx    \undefined \def \showISBNx     #1{\unskip}     \fi
\ifx \showISBNxiii \undefined \def \showISBNxiii  #1{\unskip}     \fi
\ifx \showISSN     \undefined \def \showISSN      #1{\unskip}     \fi
\ifx \showLCCN     \undefined \def \showLCCN      #1{\unskip}     \fi
\ifx \shownote     \undefined \def \shownote      #1{#1}          \fi
\ifx \showarticletitle \undefined \def \showarticletitle #1{#1}   \fi
\ifx \showURL      \undefined \def \showURL       {\relax}        \fi
\providecommand\bibfield[2]{#2}
\providecommand\bibinfo[2]{#2}
\providecommand\natexlab[1]{#1}
\providecommand\showeprint[2][]{arXiv:#2}

\bibitem[Bain et~al\mbox{.}(2021)]%
        {webvid}
\bibfield{author}{\bibinfo{person}{Max Bain}, \bibinfo{person}{Arsha Nagrani},
  \bibinfo{person}{G{\"{u}}l Varol}, {and} \bibinfo{person}{Andrew Zisserman}.}
  \bibinfo{year}{2021}\natexlab{}.
\newblock \showarticletitle{Frozen in Time: {A} Joint Video and Image Encoder
  for End-to-End Retrieval}. In \bibinfo{booktitle}{\emph{ICCV}}.
\newblock


\bibitem[Blattmann et~al\mbox{.}(2023)]%
        {VideoLDM}
\bibfield{author}{\bibinfo{person}{Andreas Blattmann}, \bibinfo{person}{Robin
  Rombach}, \bibinfo{person}{Huan Ling}, \bibinfo{person}{Tim Dockhorn},
  \bibinfo{person}{Seung~Wook Kim}, \bibinfo{person}{Sanja Fidler}, {and}
  \bibinfo{person}{Karsten Kreis}.} \bibinfo{year}{2023}\natexlab{}.
\newblock \showarticletitle{Align your Latents: High-Resolution Video Synthesis
  with Latent Diffusion Models}. In \bibinfo{booktitle}{\emph{CVPR}}.
\newblock


\bibitem[Koepke et~al\mbox{.}(2022)]%
        {Koepke21}
\bibfield{author}{\bibinfo{person}{A.~Sophia Koepke},
  \bibinfo{person}{Andreea{-}Maria Oncescu}, \bibinfo{person}{Jo{\~{a}}o~F.
  Henriques}, \bibinfo{person}{Zeynep Akata}, {and} \bibinfo{person}{Samuel
  Albanie}.} \bibinfo{year}{2022}\natexlab{}.
\newblock \showarticletitle{Audio Retrieval with Natural Language Queries: {A}
  Benchmark Study}.
\newblock \bibinfo{journal}{\emph{IEEE TMM}} (\bibinfo{year}{2022}).
\newblock


\bibitem[Li et~al\mbox{.}(2023)]%
        {AMT}
\bibfield{author}{\bibinfo{person}{Zhen Li}, \bibinfo{person}{Zuo{-}Liang Zhu},
  \bibinfo{person}{Linghao Han}, \bibinfo{person}{Qibin Hou},
  \bibinfo{person}{Chun{-}Le Guo}, {and} \bibinfo{person}{Ming{-}Ming Cheng}.}
  \bibinfo{year}{2023}\natexlab{}.
\newblock \showarticletitle{{AMT:} All-Pairs Multi-Field Transforms for
  Efficient Frame Interpolation}. In \bibinfo{booktitle}{\emph{CVPR}}.
\newblock


\bibitem[Liu et~al\mbox{.}(2022)]%
        {TTVFI}
\bibfield{author}{\bibinfo{person}{Chengxu Liu}, \bibinfo{person}{Huan Yang},
  \bibinfo{person}{Jianlong Fu}, {and} \bibinfo{person}{Xueming Qian}.}
  \bibinfo{year}{2022}\natexlab{}.
\newblock \showarticletitle{{TTVFI:} Learning Trajectory-Aware Transformer for
  Video Frame Interpolation}.
\newblock \bibinfo{journal}{\emph{CoRR}}  \bibinfo{volume}{arXiv}
  (\bibinfo{year}{2022}).
\newblock


\bibitem[Liu et~al\mbox{.}(2023)]%
        {svgvqgan}
\bibfield{author}{\bibinfo{person}{Jiawei Liu}, \bibinfo{person}{Weining Wang},
  \bibinfo{person}{Sihan Chen}, \bibinfo{person}{Xinxin Zhu}, {and}
  \bibinfo{person}{Jing Liu}.} \bibinfo{year}{2023}\natexlab{}.
\newblock \showarticletitle{Sounding Video Generator: {A} Unified Framework for
  Text-guided Sounding Video Generation}.
\newblock \bibinfo{journal}{\emph{arXiv}} (\bibinfo{year}{2023}).
\newblock


\bibitem[Oncescu et~al\mbox{.}(2021)]%
        {OncescuKHAA21}
\bibfield{author}{\bibinfo{person}{Andreea{-}Maria Oncescu},
  \bibinfo{person}{A.~Sophia Koepke}, \bibinfo{person}{Jo{\~{a}}o~F.
  Henriques}, \bibinfo{person}{Zeynep Akata}, {and} \bibinfo{person}{Samuel
  Albanie}.} \bibinfo{year}{2021}\natexlab{}.
\newblock \showarticletitle{Audio Retrieval with Natural Language Queries}. In
  \bibinfo{booktitle}{\emph{Interspeech}}.
\newblock


\bibitem[Rombach et~al\mbox{.}(2022)]%
        {latentdiffusion}
\bibfield{author}{\bibinfo{person}{Robin Rombach}, \bibinfo{person}{Andreas
  Blattmann}, \bibinfo{person}{Dominik Lorenz}, \bibinfo{person}{Patrick
  Esser}, {and} \bibinfo{person}{Bj{\"{o}}rn Ommer}.}
  \bibinfo{year}{2022}\natexlab{}.
\newblock \showarticletitle{High-Resolution Image Synthesis with Latent
  Diffusion Models}. In \bibinfo{booktitle}{\emph{CVPR}}.
\newblock


\bibitem[Ronneberger et~al\mbox{.}(2015)]%
        {unet}
\bibfield{author}{\bibinfo{person}{Olaf Ronneberger}, \bibinfo{person}{Philipp
  Fischer}, {and} \bibinfo{person}{Thomas Brox}.}
  \bibinfo{year}{2015}\natexlab{}.
\newblock \showarticletitle{U-Net: Convolutional Networks for Biomedical Image
  Segmentation}. In \bibinfo{booktitle}{\emph{MICCAI}}.
\newblock


\bibitem[Ruan et~al\mbox{.}(2023)]%
        {MMDiffusion}
\bibfield{author}{\bibinfo{person}{Ludan Ruan}, \bibinfo{person}{Yiyang Ma},
  \bibinfo{person}{Huan Yang}, \bibinfo{person}{Huiguo He},
  \bibinfo{person}{Bei Liu}, \bibinfo{person}{Jianlong Fu},
  \bibinfo{person}{Nicholas~Jing Yuan}, \bibinfo{person}{Qin Jin}, {and}
  \bibinfo{person}{Baining Guo}.} \bibinfo{year}{2023}\natexlab{}.
\newblock \showarticletitle{{MM-Diffusion}: Learning Multi-Modal Diffusion
  Models for Joint Audio and Video Generation}.
\newblock  (\bibinfo{year}{2023}).
\newblock


\bibitem[Singer et~al\mbox{.}(2022)]%
        {MakeAVideo}
\bibfield{author}{\bibinfo{person}{Uriel Singer}, \bibinfo{person}{Adam
  Polyak}, \bibinfo{person}{Thomas Hayes}, \bibinfo{person}{Xi Yin},
  \bibinfo{person}{Jie An}, \bibinfo{person}{Songyang Zhang},
  \bibinfo{person}{Qiyuan Hu}, \bibinfo{person}{Harry Yang},
  \bibinfo{person}{Oron Ashual}, \bibinfo{person}{Oran Gafni}, {et~al\mbox{.}}}
  \bibinfo{year}{2022}\natexlab{}.
\newblock \showarticletitle{{Make-A-Video}: Text-to-video generation without
  text-video data}.
\newblock \bibinfo{journal}{\emph{arXiv}} (\bibinfo{year}{2022}).
\newblock


\bibitem[Stability-AI(2022)]%
        {stablediffusion}
\bibfield{author}{\bibinfo{person}{Stability-AI}.}
  \bibinfo{year}{2022}\natexlab{}.
\newblock \bibinfo{title}{Stable Diffusion}.
\newblock
\newblock
\newblock
\shownote{\url{https://github.com/Stability-AI/StableDiffusion}}.


\bibitem[Vaswani et~al\mbox{.}(2017)]%
        {transformer}
\bibfield{author}{\bibinfo{person}{Ashish Vaswani}, \bibinfo{person}{Noam
  Shazeer}, \bibinfo{person}{Niki Parmar}, \bibinfo{person}{Jakob Uszkoreit},
  \bibinfo{person}{Llion Jones}, \bibinfo{person}{Aidan~N. Gomez},
  \bibinfo{person}{Lukasz Kaiser}, {and} \bibinfo{person}{Illia Polosukhin}.}
  \bibinfo{year}{2017}\natexlab{}.
\newblock \showarticletitle{Attention is All you Need}. In
  \bibinfo{booktitle}{\emph{NeurIPS}}.
\newblock


\bibitem[Wang et~al\mbox{.}(2023a)]%
        {bark}
\bibfield{author}{\bibinfo{person}{Chengyi Wang}, \bibinfo{person}{Sanyuan
  Chen}, \bibinfo{person}{Yu Wu}, \bibinfo{person}{Ziqiang Zhang},
  \bibinfo{person}{Long Zhou}, \bibinfo{person}{Shujie Liu},
  \bibinfo{person}{Zhuo Chen}, \bibinfo{person}{Yanqing Liu},
  \bibinfo{person}{Huaming Wang}, \bibinfo{person}{Jinyu Li},
  \bibinfo{person}{Lei He}, \bibinfo{person}{Sheng Zhao}, {and}
  \bibinfo{person}{Furu Wei}.} \bibinfo{year}{2023}\natexlab{a}.
\newblock \showarticletitle{Neural Codec Language Models are Zero-Shot Text to
  Speech Synthesizers}.
\newblock \bibinfo{journal}{\emph{arXiv}} (\bibinfo{year}{2023}).
\newblock


\bibitem[Wang et~al\mbox{.}(2023b)]%
        {VideoFactory}
\bibfield{author}{\bibinfo{person}{Wenjing Wang}, \bibinfo{person}{Huan Yang},
  \bibinfo{person}{Zixi Tuo}, \bibinfo{person}{Huiguo He},
  \bibinfo{person}{Junchen Zhu}, \bibinfo{person}{Jianlong Fu}, {and}
  \bibinfo{person}{Jiaying Liu}.} \bibinfo{year}{2023}\natexlab{b}.
\newblock \showarticletitle{VideoFactory: Swap Attention in Spatiotemporal
  Diffusions for Text-to-Video Generation}.
\newblock \bibinfo{journal}{\emph{arXiv}} (\bibinfo{year}{2023}).
\newblock


\bibitem[Zhu et~al\mbox{.}(2023)]%
        {moviefactory}
\bibfield{author}{\bibinfo{person}{Junchen Zhu}, \bibinfo{person}{Huan Yang},
  \bibinfo{person}{Huiguo He}, \bibinfo{person}{Wenjing Wang},
  \bibinfo{person}{Zixi Tuo}, \bibinfo{person}{Wen{-}Huang Cheng},
  \bibinfo{person}{Lianli Gao}, \bibinfo{person}{Jingkuan Song}, {and}
  \bibinfo{person}{Jianlong Fu}.} \bibinfo{year}{2023}\natexlab{}.
\newblock \showarticletitle{MovieFactory: Automatic Movie Creation from Text
  using Large Generative Models for Language and Images}. In
  \bibinfo{booktitle}{\emph{ACM MM BNI}}.
\newblock


\end{thebibliography}

\end{document}